\documentclass[journal]{IEEEtran}
\usepackage{graphicx}
\usepackage{subfig}
\usepackage{graphicx}
\usepackage{booktabs}
\usepackage{comment}
\usepackage{color}
\graphicspath{ {images/} }
\usepackage{subfig}
\begin{document}
\title{Semi-Automatic Algorithm for Breast MRI Lesion Segmentation Using Marker-Controlled Watershed Transformation}

\author{Sulaiman Vesal,
        Andres Diaz-Pinto,
        Nishant Ravikumar, 
        Stephan Ellmann,
        Amirabbas Davari\\
        and~Andreas Maier% <-this % stops a space
\thanks{Manuscript received on November 8, 2017. This work was supported in part by Emerging Field Initiative (EFI) project of Friedrich-Alexander-University Erlangen-Nuremberg.}% <-this % stops a space
\thanks{S. Vesal, N.Ravikumar, and A. Maier  are with the Pattern Recognition Lab, Friedrich-Alexander-University Erlangen-Nuremberg, Erlangen, Germany (telephone: +49 152 10228542, e-mail: \{sulaiman.vesal, nishant.kumar, amir.davari, andreas.maier\}@fau.de).}%
\thanks{A. Diaz-Pinto is with Instituto de Investigaci\'on e Innovaci\'on en Bioingenier\'ia, Universitat Polit\`ecnica de Val\`encia, Valencia, Spain (e-mail: andiapin@upv.es).}%
\thanks{S. Ellmann is with Radiologisches Institut, Universit\"atsklinikum Erlangen, Erlangen, Germany (e-mail: stephan.ellmann@uk-erlangen.de).}%
}

% \IEEEoverridecommandlockouts
% \IEEEpubid{\centering \makebox[\columnwidth]{978-1-5386-2282-7/17/\$31.00~
% \copyright2017~IEEE}\hspace{\columnsep}}

\twocolumn[
\begin{@twocolumnfalse}

\maketitle

\begin{abstract}
Magnetic resonance imaging (MRI) is an effective imaging modality for identifying and localizing breast lesions in women. Accurate and precise lesion segmentation using a computer-aided-diagnosis (CAD) system, is a crucial step in evaluating tumor volume and in the quantification of tumor characteristics. However, this is a challenging task, since breast lesions have sophisticated shape, topological structure, and high variance in their intensity distribution across patients. In this paper, we propose a novel marker-controlled watershed transformation-based approach, which uses the brightest pixels in a region of interest (determined by experts) as markers to overcome this challenge, and accurately segment lesions in breast MRI. The proposed approach was evaluated on 106 lesions, which includes 64 malignant and 42 benign cases. Segmentation results were quantified by comparison with ground truth labels, using the  Dice similarity coefficient (DSC) and Jaccard index (JI) metrics. The proposed method achieved an average dice coefficient of 0.7808 $\pm$ 0.1729 and Jaccard index of 0.6704 $\pm$ 0.2167. These results illustrate that the proposed method shows promise for future work related to the segmentation and classification of benign and malignant breast lesions. \\

\end{abstract}
\end{@twocolumnfalse}
]
{\renewcommand{\thefootnote}%
{\fnsymbol{footnote}}
\footnotetext[1]{S. Vesal, N.Ravikumar, and A. Maier  are with the Pattern Recognition Lab, Friedrich-Alexander-University Erlangen-Nuremberg, Germany}
\footnotetext[1]{A. Diaz-Pinto is with Instituto de Investigaci\'on e Innovaci\'on en Bioingenier\'ia, Universitat Polit\`ecnica de Val\`encia, Spain}
\footnotetext[1]{S. Ellmann is with Radiologisches Institut, Universit\"atsklinikum Erlangen, Germany}
}
\markboth{2017 IEEE Nuclear Science Symposium and Medical Imaging Conference}{}
%\begin{IEEEkeywords}
%IEEEtran, journal, \LaTeX, paper, template.
%\end{IEEEkeywords}

\section{Introduction}
In recent years, breast cancer has been one of the major causes of death in women \cite{Nishikawa}\cite{Hu}. Clinical evidence indicates that early detection and treatment can significantly reduce the mortality of breast cancer. Various medical imaging modalities play a crucial role in detection, diagnosis and treatment planning of breast cancer. Among these modalities,  magnetic resonance imaging (MRI) is used primarily for women at high risk of developing cancer, to identify lesions missed in the mammogram, and typically require accurate lesion segmentation as an initial step in breast MRI-based CAD systems \cite{McClymont}\cite{prior}. However, accurate segmentation of lesions is a challenging problem for many reasons such as, lesions have considerable variation in terms of shape, having an overlapping area with normal tissues which is difficult to differentiate and distribution of intensities in the lesions are very high, etc. 

However,  many researchers have explored various semi-automated and automated approaches to address these challenges. Normally, breast tumor segmentation methods fall into main three categories which are model-based, threshold-based and region growing based methods. Chen et al. \cite{Chen} proposed a fuzzy c-means (FCM) based method. Szabó et al. \cite{szab} performed a pixel-by-pixel lesion segmentation through an artificial neural network based on the time intensity curve. Shanon et al. \cite{Shanon} proposed spectral embedding based active contour to improve image representation for both boundary and region-based segmentation. However, most of these methods require post-processing through morphological operations such as dilation and erosion, to obtain a smooth and continuous lesion boundary. Xu et al. \cite{Xu} proposed a watershed transformation method for mammogram mass segmentation using markers. Their method involved identifying the rough region of the lesion by a combined template matching and thresholding approach as an initial step, followed by identifying internal markers by computing the distance transform from the rough segmentation, and external markers through morphological dilation.\\

In this paper, we propose a novel marker-controlled watershed transformation (MCWT) for the task of 2D breast lesion segmentation in MRI slices. This technique is more robust and simpler in marker selection process, in comparison to conventional methods which used complex features such as template matching and Gaussian mixture model \cite{Xu}\cite{Cui}. Section \ref{sec:methods} describes the data acquisition, the proposed method and evaluation results and our work will be concluded in Section \ref{sec:conclude}. 

\section{Methods}
\label{sec:methods}
\subsection{Data Acquisition}
MR images for this study were acquired on 1.5 T scanners Magnetom Avanto and 3.0 T  Magnetom Verio, Siemens Healthineers, Erlangen, Germany, with dedicated breast array coils and the patient in a prone position. The contrast media was applied into the cubital vein after the first of six dynamic acquisitions with a flow of 1.0 mL/sec chased by a 20 mL saline flush. One hundred and six lesions were identified from a representative set of DCE-MRI exams from 80 female patients by two expert radiologists who have 7 years of experience in evaluation of clinical findings. The mean patient age was 50 $\pm$ 13  and based on histopathologically 42 of the lesions were diagnosed as benign and the remaining 64 as malignant. 

\subsection{Watershed Transformation}
The watershed transformation is a widely used method for image segmentation based on mathematical morphology \cite{Beucher}. In this method, the image is considered as a topographic relief or landscape where the gray level of each pixel corresponds to a physical elevation. The landscape is immersed in a lake,  with holes pierced in local minima, and the catchment areas/basins are filled up with water/labels starting at these local minima. The points where water floods from different basins, a dam/border is built. The process terminates when the water level reaches a maximum. Consequently, the landscape is divided into several regions separated by dams which are called watershed lines or basically watersheds \cite{water}.

The main drawback of the conventional watershed transformation is over-segmentation which is caused due to the presence of many local minima in the gradient image. To decrease the effect of severe over-segmentation,  we determine internal and external markers to guide the watershed algorithm.  Each marker is considered to be part of a specific watershed region and after segmentation, the boundaries of the regions are arranged to separate each object of interest from the background.

The proposed method starts with selecting a particular slice of breast MRI volume, performed by an expert radiologist, such that it contains at least one lesion. The slice image is normally the subtraction of pre-contrast and post-contrast images. The ground truth segmentation provided by expert radiologists and ROI is drawn around the lesion manually. As a pre-processing step, We applied contrast limited adaptive histogram equalization (CLAHE) \cite{Clahe} on that particular slice globally to improve lesion contrast. Then we computed the morphological gradient of the image, which is the point-wise difference between a unitary dilation and erosion.

In MR images, tumor regions are normally brighter and have more uniform intensity than the neighbouring healthy tissue. Based on this fact, we determined the internal and external markers by sorting out the pixel values in ROIs in descending order and chose $n$ pixels with maximum intensity values as markers. After selecting the markers the normal watershed transformation is applied on the ROIs image which is shown in Fig 1 Finally, a binary mask is generated based on watershed output regions. However, we identified the optimal number of markers based on segmentation accuracy evaluated using Dice and Jaccard.

\begin{figure}[t]
    \subfloat[\label{fig:test1}]
    {\includegraphics[width=2.5cm,height=2.5cm]{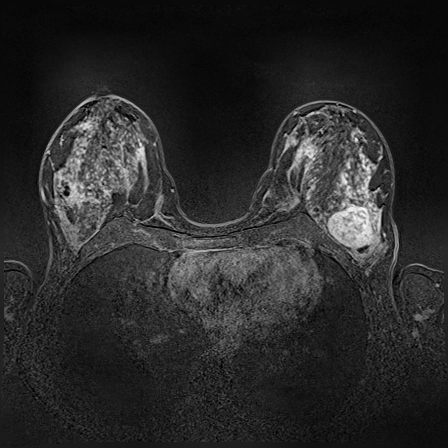}}\hfill
    \subfloat[\label{fig:test2}]
    {\includegraphics[width=2.5cm,height=2.5cm]{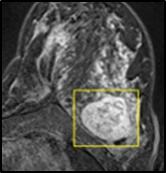}}\hfill
    \subfloat[\label{fig:test3}]
    {\includegraphics[width=2.5cm,height=2.5cm]{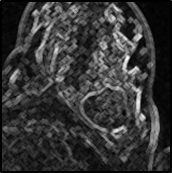}}\hfill
    \subfloat[\label{fig:test4}]
    {\includegraphics[width=2.5cm,height=2.5cm]{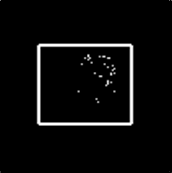}}\hfill
    \subfloat[\label{fig:test5}]
    {\includegraphics[width=2.5cm,height=2.5cm]{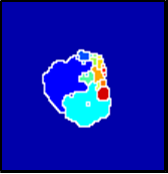}}\hfill
    \subfloat[\label{fig:test6}]
    {\includegraphics[width=2.5cm,height=2.5cm]{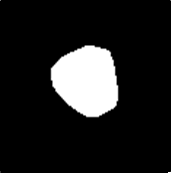}}
    \caption{The segmentation pipeline.  (a) Subtraction slice with a benign lesion.  (b) Contrast enhancement using CLAHE and ROI is drawn. (c) Image gradient. (d) The highest pixel intensities are selected as markers. (e) Watershed transformation applied. (f) Segmentation mask generated based on watershed.}
    \label{imgss}
    \centering
\end{figure}

\subsection{Evaluation and Results}

\begin{figure}[h!]
    \includegraphics[width=8.8cm]{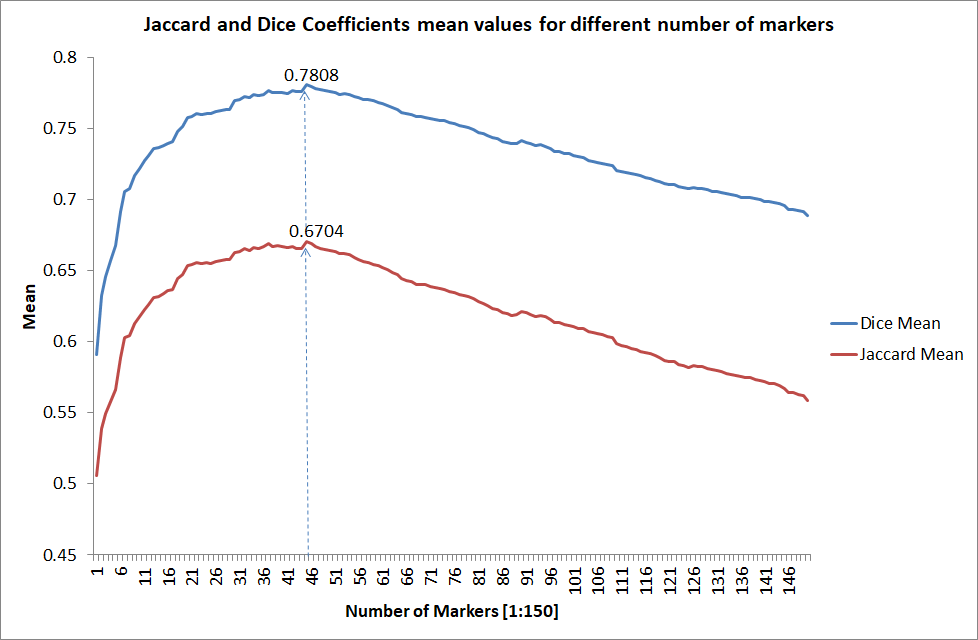}
    \caption{The mean of total lesions for Jaccard and Dice coefficients with different number of markers. 
    }
\label{graph}
    \centering
\end{figure}

We tested the algorithm by varying the number of markers between $1$ and $150$. Fig~\ref{graph} describes the segmentation results obtained using different numbers of markers. This plot indicates that $45$ markers were found to be optimal using this segmentation approach, yielding satisfactory results. 

\begin{figure}
%    \subfloat[Worst Case\label{fig:test1}]
    {\includegraphics[width=4.2cm,height=4.2cm]{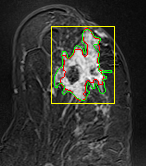}}\hfill
%    \subfloat[Average Case\label{fig:test2}]
    {\includegraphics[width=4.2cm,height=4.2cm]{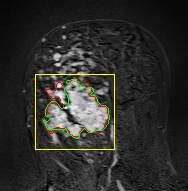}}\hfill \\
%    \subfloat[Best Case\label{fig:test3}]\\
    {\includegraphics[width=4.2cm,height=4.2cm]{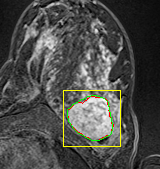}}\hfill
    {\includegraphics[width=4.2cm,height=4.2cm]{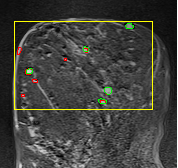}}\hfill
    
    \caption{Segmentation results on 4 breast lesion cases. The top two images show malignant lesions and the two down show benign lesions. The yellow rectangle represents the ROIs, ground truth segmentation represented in green line and the result of proposed method shown in the red line.}
    \label{imgs}
\end{figure}

To evaluate the performance of the proposed method and compare the segmentation results with their corresponding ground truth labels, two well-known similarity metrics are used: (1) the Dice similarity coefficient (DSC), which measures the overlap between two segmentation masks and is sensitive to the lesion size, (2) the Jaccard index (JI), which denotes the average distance between two segmentations \cite{prior}. The metrics are defined as:

\begin{equation}
DSC (A,B) = \frac{2|A \cap B|}{|A|+|B|}
\end{equation}

\begin{equation}
JI (A,B) = \frac{|A \cup B|}{|A \cap B|}
\end{equation}

where $A$ refers to the ROIs segmented by our algorithm and $B$ is tumor area as determined by manual segmentation. Table 1 summarizes the segmentation accuracy achieved using the proposed method for all 106 cases. The average dice coefficient was found to be 0.78$\pm$0.17 and average Jaccard index was 0.67$\pm$0.21. Fig~\ref{imgs} demonstrate four sample segmentation outputs which are overlaid on manual segmentations provided by two radiologists. It can be seen, that the proposed method could accurately segment the lesions with some marginal errors for medium to large tumors. However, for cases comprising disjoint lesions, the method failed to segment all small lesions and in some cases incorrectly labeled healthy tissue as lesions. This is because in some cases there is a high degree of overlap in the intensity distributions of healthy breast tissue and lesions, and the ROI drawn by the radiologist is very large in the case of disjoint lesions, in order to cover the entire area over which multiple lesions are distributed.

\renewcommand{\arraystretch}{1.5}
\begin{table}[h!]
\centering
\caption{DSC and JI results $(mean\pm Std)$ for 106 lesions.}
\label{res}
\begin{tabular}{c|c|c}
\hline
 Method& DSC & JI  \\ \hline
 MCWT& 0.7808$\pm$0.1729 & 0.6704$\pm$0.2167 \\ \hline
\end{tabular}
\end{table}

\section{Discussion and Conclusion}
\label{sec:conclude}
Segmentation of breast lesions in MR images has been tackled previously in various studies, however, very few have employed the marker-controlled watershed transformation approach for this purpose. In this paper, we proposed a novel marker-controlled watershed transformation approach by selecting the brightest pixels as markers in the ROIs. In terms of complexity, this method is simpler and robust in comparison to conventional marker-based watershed methods which used complex features to determine external and internal markers. However, the diversity of lesion shapes and the presence of multiple disjoint lesions distributed across the breast proved challenging, resulting in low DSC and JI scores in some cases. These preliminary results are encouraging for the application of the proposed approach, as a preprocessing step for subsequent classification of MRI lesions. Manually-created ground truth images are intrinsically subjective and creating such reference images for large data sets is a time-consuming process. In subsequent studies, we will look to extend our proposed 2D watershed algorithm to 3D and combine it with a lesion detection and classification technique, to establish a complete computer-aided-diagnosis system, with minimum manual intervention. 

\section*{Acknowledgment}
The authors gratefully acknowledge the support of Emerging Field Intuitive (EFI) project. We also thank our colleagues from Universit\"atsklinikum Erlangen who provided the dataset and insight that greatly assisted this research.

\bibliographystyle{IEEEtran}

\bibliography{biblo}

\end{document}